\documentclass[runningheads]{llncs}

\usepackage[T1]{fontenc}
\usepackage[utf8]{inputenc}
\usepackage{graphicx}
\usepackage{amsmath, amssymb}
\usepackage{ragged2e}
\usepackage{tabularx}
\usepackage{array}
\usepackage{booktabs} 

\usepackage{pgfplots}
\pgfplotsset{compat=1.18}

\usepackage{xcolor} 
\usepackage{listings}
\usepackage{tikz}   
\usetikzlibrary{shapes.arrows, shadows} 

\usepackage[most]{tcolorbox} 

\definecolor{codegray}{rgb}{0.95,0.95,0.95}
\definecolor{codeblue}{rgb}{0.1,0.3,0.6}
\definecolor{codepurple}{rgb}{0.5,0,0.35}
\definecolor{jsonkey}{rgb}{0.1, 0.4, 0.5}
\definecolor{jsonstr}{rgb}{0.3, 0.1, 0.1}
\definecolor{backcolour}{rgb}{0.98, 0.98, 0.98}

\tcbset{
    promptstyle/.style={
        enhanced,
        breakable,
        colback=gray!5!white,
        colframe=black!75,
        boxrule=0.8pt,
        arc=3pt,
        left=6pt, right=6pt, top=4pt, bottom=4pt,
        fonttitle=\bfseries\small,
        fontupper=\small,
        coltitle=white,
        attach boxed title to top left={yshift=-2.5mm, xshift=4mm},
        boxed title style={colback=black!75, arc=2pt, boxrule=0pt}
    }
}

\usepackage{url}

\usepackage{tabularx}
\usepackage{booktabs}
\usepackage{xcolor}
\usepackage{enumitem} 
\usepackage{caption}  

\begin{document}

\title{TopoChunker: Topology-Aware Agentic Document Chunking Framework}
\titlerunning{TopoChunker} 

\author{Xiaoyu Liu}
\authorrunning{Xiaoyu Liu}
\institute{Independent Researcher, Beijing, China \\
\email{liushifu986938279@gmail.com}}

\maketitle

\begin{abstract}
Current document chunking methods for Retrieval-Augmented Generation (RAG) typically linearize text. This forced linearization strips away intrinsic topological hierarchies, creating ``semantic fragmentation'' that degrades downstream retrieval quality. In this paper, we propose TopoChunker, an agentic framework that maps heterogeneous documents onto a Structured Intermediate Representation (SIR) to explicitly preserve cross-segment dependencies. To balance structural fidelity with computational cost, TopoChunker employs a dual-agent architecture. An Inspector Agent dynamically routes documents through cost-optimized extraction paths, while a Refiner Agent performs capacity auditing and topological context disambiguation to reconstruct hierarchical lineage. Evaluated on unstructured narratives (GutenQA) and complex reports (GovReport), TopoChunker demonstrates state-of-the-art performance. It establishes new benchmarks across both domains, achieving an 83.26\% Recall@3 on highly structured data and outperforming advanced generative and dynamic routing baselines by up to 8.0\% in absolute generation accuracy.The source code, prompts, and synthetic datasets will be publicly released upon acceptance. Our code is available at: \url{https://github.com/liushifu12138/TopoChunker}

\keywords{Retrieval-Augmented Generation \and Document Chunking \and Multi-Agent Systems \and Structured Intermediate Representation.}
\end{abstract}

\section{Introduction}

Retrieval-Augmented Generation (RAG) mitigates Large Language Model (LLM) hallucinations by grounding outputs in external knowledge. Despite advances in long-context LLMs, fine-grained retrieval remains essential to minimize inference costs and avoid the ``Lost in the Middle'' phenomenon. Consequently, document chunking acts as a foundational filter, directly determining the upper bound of downstream retrieval quality.

However, conventional and emerging LLM-based chunkers (e.g., LumberChunker~\cite{duarte2024}, Mixture-of-Chunkers) predominantly rely on a rigid ``text-to-text'' paradigm. Flattening complex, multi-modal documents into linear streams destroys inherent hierarchical topology, creating ``semantic islands'' that isolate terminology and cause anaphoric ambiguity. Furthermore, fully generative chunkers apply expensive LLM reasoning uniformly, resulting in redundant token expenditures on structurally simple content, while purely granularity-aware routers fail to preserve explicit physical relationships.

To address these limitations, we propose TopoChunker, a topology-aware framework for structural-fidelity chunking. Instead of linear pipelines, TopoChunker maps heterogeneous documents onto a Structured Intermediate Representation (SIR). Driven by a collaborative dual-agent architecture---an Inspector Agent for routing and a Refiner Agent for semantic auditing---the framework systematically transforms fragmented raw text into contextually self-contained RAG inputs.

Our primary contributions are summarized as follows:
\begin{itemize}
    \item \textbf{Adaptive Routing via Inspector Agent:} We introduce a dynamic routing mechanism that probes document complexity to optimize the cost-fidelity trade-off. By selectively applying rule-based slicing for standard layouts and Vision-Language Models (VLMs) for complex structures, it effectively prevents uniform LLM overhead.
    \item \textbf{Topological Modeling \& Deterministic Execution:} We propose the SIR protocol to decouple physical parsing from semantic reasoning. By integrating pointer networks and Atomicity Locking, this in-memory graph ensures ``zero-hallucination'' extraction, explicitly preserving logical hierarchies and indivisible units (e.g., cross-page tables).
    \item \textbf{Context Disambiguation via Refiner Agent:} We design a diagnostic Refiner Agent to resolve contextual voids (e.g., dangling pronouns) and generate self-contained chunks. TopoChunker establishes new state-of-the-art benchmarks (e.g., 83.26\% Recall@3 on structured data). It definitively outperforms both generative baselines like LumberChunker~\cite{duarte2024} (+8.0\% absolute accuracy on unstructured narratives) and dynamic routers like MoC~\cite{zhao2025moc} (+3.0\% on complex reports). Furthermore, instead of blindly expending tokens, TopoChunker maximizes semantic ROI---slashing redundant reasoning overhead on structured layouts by 26.7\% while strategically reallocating budgets to resolve unstructured ambiguity.
\end{itemize}

\section{Related Work}

\subsection{Granularity Evolution: From Heuristic Rules to Semantic Sensitivity}
Document chunking has shifted from rigid heuristics—which frequently cause semantic misalignment—to semantic-driven methods. Approaches like Semantic Chunker~\cite{xiao2024cpack}, Meta-Chunking~\cite{zhao2024}, and LumberChunker~\cite{duarte2024} leverage embeddings or LLM reasoning for boundary prediction. However, two critical problems persist: purely generative methods lack deterministic physical grounding (risking boundary hallucinations), and oversized chunks suffer from ``feature dilution,'' which degrades retrieval. TopoChunker resolves this using a Structured Intermediate Representation (SIR) for precise, hallucination-free granularity adjustment.

\subsection{Hierarchical Representation and Context Augmentation in RAG}
Current hierarchical RAG frameworks often fail to capture deep physical topologies. For instance, Late Chunking~\cite{gunther2024} outputs flat lists, while MoC~\cite{zhao2025moc} remains topology-agnostic. Furthermore, Parent-Child Indexing frequently creates ``semantic islands,'' and many advanced systems incur high I/O overhead by relying on external graph databases. TopoChunker overcomes these limitations via a lightweight, in-memory SIR. By dynamically injecting Topological Lineage, it explicitly restores contextual dependencies without requiring external databases.

\subsection{Agentic Workflows and Adaptive Diagnostics for Document Intelligence}
Autonomous agents enable ``reasoning-before-action'' but face significant efficiency challenges. Models like PIC~\cite{wang2025segmentation} lack spatial awareness, while Vision-Language Models (VLMs) incur prohibitive computational costs. Additionally, emerging agentic chunkers waste tokens by uniformly applying expensive LLM reasoning across all standard structural elements. TopoChunker mitigates this with a ``Diagnosis-Execution-Audit'' paradigm. It employs an Inspector Agent to minimize redundant processing and a ReAct-driven Refiner Agent to strictly allocate generative reasoning only to high-complexity data, thereby maximizing preprocessing efficiency.

\begin{figure*}[t]
    \centering
    \includegraphics[width=\textwidth]{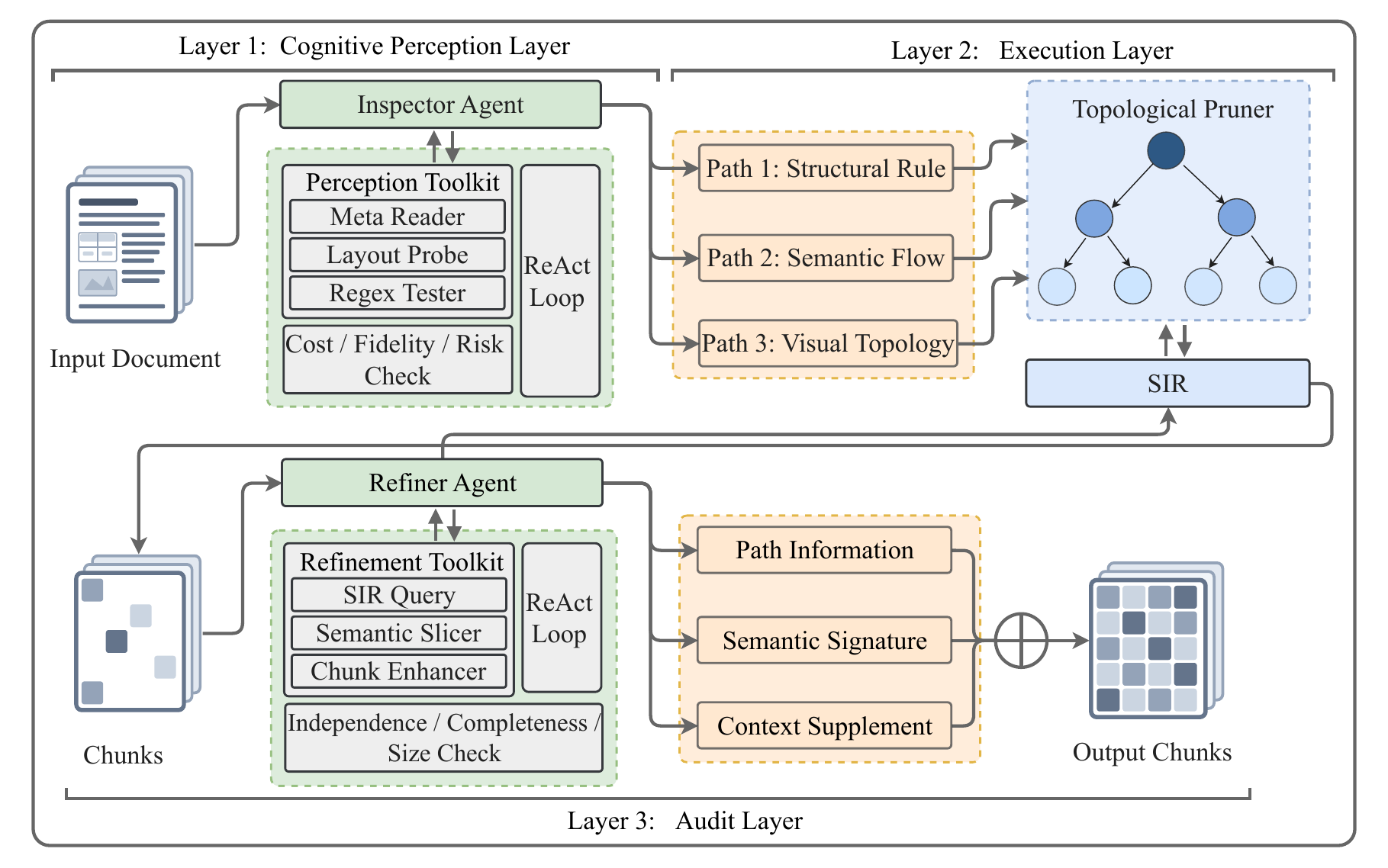}
    \caption{The overall architecture of TopoChunker. The dual-agent framework operates across three layers: Cognitive Perception (Inspector Agent for adaptive routing), Execution (Topological Pruner and SIR construction), and Audit (Refiner Agent for context disambiguation). These layers interact via a shared Structured Intermediate Representation (SIR) to form a closed-loop Diagnosis-Execution-Audit pipeline.}
    \label{fig:architecture}
\end{figure*}

\section{Method}

\subsection{Cognitive Perception and Adaptive Routing}
\label{sec:perception}

TopoChunker employs a three-stage agentic architecture driven by reasoning-focused LLMs . The process initiates in the Cognitive Perception Layer, where an Inspector Agent evaluates the trade-off between information loss and computational cost.

\subsubsection{Active Probing and Path Selection.}
Instead of consuming raw documents entirely, the Inspector Agent operates within a ReAct framework to sparsely sample documents via lightweight APIs. This constructs a compact topology fingerprint $\vec{f}$ (under 200 tokens). By evaluating $\vec{f}$ with a JSON-constrained prompt loop, the Inspector avoids full-text processing and unbounded CoT costs, dynamically routing heterogeneous documents into three paths:

\begin{description}
    \item[Path 1: Structural Rule.] For digital documents with clear hierarchies. It utilizes pointer networks and mapping tables to dynamically intercept logical patterns, strictly avoiding generative extraction for ``zero-hallucination'' parsing.
    \item[Path 2: Semantic Flow.] For unstructured narratives. The Inspector routes these to an LLM-based \texttt{Semantic\_Slicer}, which evaluates the flat text to prioritize semantic continuity.
    \item[Path 3: Visual Topology.] For scanned or complex layouts. This pipeline leverages layout-aware VLMs (e.g., MinerU~\cite{mineru2024}) to physically parse documents, explicitly preserving coordinates and the integrity of non-textual elements.
\end{description}

\subsection{Topological Modeling and SIR Construction}
\label{sec:modeling}

Routing outputs are unified into Markdown. To mitigate ``contextual fragmentation,'' the Topological Pruner uses a stack-based algorithm to construct a hierarchical tree topology $\mathcal{T}_{SIR}$. Each node $v \in \mathcal{T}_{SIR}$ is defined by three core attributes: 1) \textbf{Topological Lineage} ($v.\mathcal{T}_{id}$) maintains a hierarchy stack of ancestor coordinates for explicit contextual inheritance; 2) \textbf{Atomicity Locking} ($v.\mathcal{A}_{lock}$) flags indivisible elements (e.g., images, tables) to prevent physical splitting; 3) \textbf{Token Audit} ($v.W_{token}$) pre-calculates the node's token weight, enabling $\mathcal{O}(1)$ capacity evaluation for downstream agents.

\subsection{Semantic Refinement and Context Assembly}
\label{sec:refinement}

To balance structural completeness and retrieval precision, the Refiner Agent manages segments via a ReAct paradigm~\cite{yao2023react} using three specialized tools:

\noindent\textbf{Capacity Auditing (\texttt{Semantic\_Slicer}):} Nodes exceeding a token threshold $\theta$ trigger the slicer. It indexes sentence boundaries and prompts a zero-shot LLM to split text precisely at topic/temporal shifts. IDs are deterministically validated to prevent hallucinations.

\noindent\textbf{Signature Generation (\texttt{Chunk\_Enhancer}):} Transforms raw segments into self-contained units by generating a concise (3-to-8 word) thematic title that abstracts the chunk's core topic for better lexical discriminability.

\noindent\textbf{Context Disambiguation (\texttt{SIR\_Query}):} Audits chunk payloads for unresolved anaphora. Upon detecting ``semantic islands,'' it triggers \texttt{SIR\_Query} to deterministically navigate $\mathcal{T}_{SIR}$ via topological pointers, retrieving ancestor context and entity definitions without external graph databases.

\section{Experiments}

\subsection{Experimental Setup}

\noindent \textbf{Datasets.} We evaluate our framework on two benchmarks: GutenQA and GovReport. Specifically, GutenQA comprises 100 public domain narrative books serving as the retrieval corpus, paired with 3,000 specific question-answer pairs for evaluation. For GovReport, we adapted it for retrieval QA by constructing a synthetic evaluation set utilizing GPT-4. We sampled 500 of these generated queries to form our final test set. The sampled QA pairs underwent a rigorous manual filtering process by two independent domain experts to verify factual grounding and resolve ambiguities, achieving high inter-annotator agreement. This curation ensured that the queries reflect realistic, high-frequency information needs of enterprise users navigating complex industrial documents.

\noindent We compare TopoChunker against five representative paradigms to evaluate its effectiveness. For a fair comparison of generative capabilities and computational overhead, the underlying reasoning model for all LLM-based approaches (including LumberChunker and our TopoChunker) is strictly standardized to DeepSeek-R1~\cite{deepseek2025r1}. 

\begin{itemize}
    \item \textbf{Fixed-Size Chunking (FC200)}: A standard heuristic baseline.
    \item \textbf{Semantic Chunker} \cite{xiao2024cpack}: A dynamic boundary method based on embedding similarity shifts.
    \item \textbf{Proposition-Level Chunking} \cite{chen-etal-2024-dense}: A granular strategy decomposing text into self-contained atomic factual statements.
    \item \textbf{LumberChunker} \cite{duarte2024}: A recent approach using LLMs to predict optimal breaks. To prevent unbounded CoT generation and ensure fair token evaluation, we explicitly constrained its prompts to enforce strict JSON boundary outputs.
    \item \textbf{Mixture-of-Chunkers (MoC)} \cite{zhao2025moc}: A recently proposed framework that dynamically routes text to specialized, granularity-aware small language models (SLMs). 
\end{itemize}

\subsection{Quantitative Evaluation}

\begin{table*}[t]
\centering
\caption{Domain-Specific RAG Performance. Performance is reported using Recall@$k$ (R@$k$) and DCG@$k$ (D@$k$). TopoChunker achieves highly competitive performance across both domains.}
\label{tab:main_results}
\begin{tabularx}{\textwidth}{@{}p{3.2cm}*{4}{>{\centering\arraybackslash}X}*{4}{>{\centering\arraybackslash}X}@{}}
\toprule
& \multicolumn{4}{c}{\textbf{GutenQA}} & \multicolumn{4}{c}{\textbf{GovReport}} \\
\cmidrule(lr){2-5} \cmidrule(lr){6-9}
& \textbf{R@3} & \textbf{D@3} & \textbf{R@5} & \textbf{D@5} & \textbf{R@3} & \textbf{D@3} & \textbf{R@5} & \textbf{D@5} \\
\midrule
Recursive (FC200) & 8.23 & 7.38 & 8.62 & 7.84 & 32.25 & 28.40 & 36.25 & 29.81 \\
Semantic Chunker & 14.49 & 12.43 & 16.56 & 13.29 & 55.37 & 48.96 & 61.75 & 51.39 \\
Proposition-Level & 47.23 & 42.27 & 53.22 & 45.73 & 66.17 & 61.36 & 73.88 & 59.44 \\
LumberChunker & 62.58 & 56.56 & 66.93 & 58.35 & 81.38 & 74.88 & 85.20 & 75.55 \\
MoC & 63.85 & \textbf{57.40} & 68.45 & 58.80 & 82.15 & 75.50 & \textbf{86.85} & 76.10 \\
\midrule
\textbf{TopoChunker} & \textbf{64.59} & 56.95 & \textbf{70.29} & \textbf{59.03} & \textbf{83.26} & \textbf{76.59} & 86.58 & \textbf{76.93} \\
\bottomrule
\end{tabularx}
\end{table*}

\subsubsection{Retrieval Performance.}
As Table~\ref{tab:main_results} indicates, TopoChunker achieves state-of-the-art retrieval accuracy across both domains. Unlike traditional methods that degrade under structural variance, TopoChunker maintains contextual integrity regardless of document complexity. On unstructured narratives (GutenQA), it outperforms advanced baselines like LumberChunker and MoC. While MoC's dynamic SLM routing is highly competitive on flat texts due to granularity optimization, TopoChunker's explicit semantic disambiguation still secures the highest R@3 (64.59\%).

Crucially, this performance advantage widens significantly on the dense hierarchies of GovReport. Although MoC is granularity-aware, its retrieval performance plateaus here because it fundamentally operates under a topology-agnostic, text-flattening paradigm. In contrast, TopoChunker avoids this bottleneck by explicitly preserving hierarchical dependencies, dominating structured retrieval (e.g., 83.26\% R@3, 76.59\% D@3). This empirical gap confirms that topological fidelity is a prerequisite for parsing complex enterprise documents.

\begin{figure}[t]
    \centering
    \includegraphics[width=0.55\linewidth]{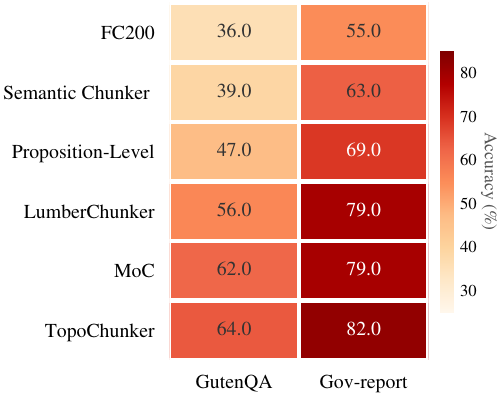}
    \caption{Generation Accuracy. A heatmap illustrating the generation accuracy of various chunking methods across the GutenQA and GovReport datasets. Darker shades indicate higher accuracy.}
    \label{fig:accuracy_heatmap}
\end{figure}

\subsubsection{Generation Accuracy.}
End-to-end RAG generation is evaluated via a manually audited, GPT-4-powered LLM-as-a-judge pipeline to ensure metric reliability. TopoChunker yields 64.0\% accuracy on unstructured GutenQA (+2.0\% over MoC) and 82.0\% on the highly structured GovReport, definitively outperforming both heuristic and advanced agentic baselines.

This advantage stems directly from our topological context disambiguation. Notably, on GovReport, the granularity-aware MoC plateaus at 79.0\%, failing to improve upon LumberChunker. This bottleneck confirms that topology-agnostic, text-flattening paradigms inherently generate ``semantic islands'' burdened by unresolved anaphora and decoupled terminology.

\subsection{Token Efficiency and CoT Overhead Analysis}

\begin{table*}[t]
\centering
\caption{Comprehensive Offline Generative LLM Token Consumption Analysis (Per 10,000 raw words $\approx$ 13,000 tokens). TopoChunker explicitly breaks down DeepSeek-R1's CoT overhead.}
\label{tab:token_cost}
\resizebox{\textwidth}{!}{%
\begin{tabular}{@{} ll cccc @{}} 
\toprule
\textbf{Method} & \textbf{Dataset (Topology Type)} & \textbf{Input Tokens} & \textbf{Output Tokens} & \textbf{CoT (\texttt{<think>})} & \textbf{Total Tokens} \\
\midrule
FC200 / Semantic Chunker & Both (Heuristic) & 0 & 0 & 0 & \textbf{0} \\
MoC (SLM 1.5B) \cite{zhao2025moc} & Both (Dynamic Routing) & $\sim$14,500 & $\sim$800 & 0 (No CoT) & \textbf{$\sim$15,300} \\
\midrule
\multicolumn{6}{l}{\textit{Generative LLM-based (DeepSeek-R1)}} \\
LumberChunker & GovReport (Structured) & 18,500 & 1,500 & 22,000 & \textbf{42,000} \\
\textbf{TopoChunker (Ours)} & GovReport (Structured $\rightarrow$ Path 1) & 22,350 & 2,450 & 6,000 & \textbf{30,800} \\
\cmidrule{1-6}
LumberChunker & GutenQA (Unstructured) & 18,500 & 1,500 & 14,000 & \textbf{34,000} \\
\textbf{TopoChunker (Ours)} & GutenQA (Unstructured $\rightarrow$ Path 2) & 34,350 & 2,900 & 8,500 & \textbf{45,750} \\
\bottomrule
\end{tabular}%
}
\end{table*}

Computational overhead is a critical bottleneck in agentic workflows utilizing reasoning models (e.g., DeepSeek-R1~\cite{deepseek2025r1}) due to extensive Chain-of-Thought (CoT) generation. By benchmarking token consumption on a standard 10,000-word ($\approx$13,000 tokens) document (Table~\ref{tab:token_cost}), we demonstrate how TopoChunker shifts the token budget from redundant boundary searching to high-ROI metadata enrichment.

\textbf{Cost Reduction on Structured Data (Path 1):} On complex documents (GovReport), generative baselines struggle to deduce the boundaries of implicit tabular structures and nested lists, triggering a massive CoT explosion ($>$22,000 \texttt{<think>} tokens). Conversely, TopoChunker routes these to Path 1, bypassing LLM-based physical slicing entirely. Its 6,000 CoT tokens are instead exclusively distributed across the Refiner Agent's multiple short-polling API calls for title generation and contextual disambiguation, cutting total consumption by \textbf{26.7\%} relative to LumberChunker.

\textbf{Accuracy-Efficiency Trade-off on Flat Narratives (Path 2):} For unstructured texts (GutenQA), TopoChunker invokes the \texttt{Semantic\_Slicer} (Path 2). Unlike baselines that rely on computationally expensive string matching, our slicer pre-indexes sentences with regex-generated IDs. This structural constraint drastically shrinks the LLM's search space, curbing CoT to 8,500 tokens compared to LumberChunker's 14,000. However, due to the dual-agent's comprehensive multi-pass reading, this rigorous processing incurs a 34.6\% total token premium. Rather than wasting tokens on raw slicing, TopoChunker explicitly reallocates this budget to construct Topological Lineage and Context Supplements, yielding a definitive \textbf{+8.0\%} absolute gain in generation accuracy.

\subsection{Module-wise Ablation Study}
\label{sec:ablation}

\begin{table}[t]
\centering
\caption{Ablation Study: Impact of Refiner and Inspector evaluated on the heterogeneous GovReport dataset.}
\label{tab:ablation}
\small 
\renewcommand{\arraystretch}{1.1} 
\begin{tabular}{l c c c} 
\toprule
\textbf{Method} & \textbf{Recall@3} & \textbf{Acc. (\%)} & \textbf{Token Cost} \\
\midrule
LumberChunker & 81.38 & 79.0 & 1.65x \\
MoC (SLM Routing) & 82.15 & 79.0 & 1.18x \\
Topo (w/o Refiner) & 76.50 & 71.0 & \textbf{0.85x} \\
\midrule
\textit{w/o Inspector (Forced Static)} & & & \\ 
\quad Rule-Only (Path 1) & 77.80 & 73.5 & 0.80x \\
\quad Semantic-Only (Path 2) & 79.20 & 76.8 & 0.95x \\
\quad VLM-Only (Path 3) & 82.85 & 81.5 & 1.88x \\
\midrule
\textbf{TopoChunker (Full)} & \textbf{83.26} & \textbf{82.0} & \textbf{1.26x} \\
\bottomrule
\end{tabular}
\end{table}

We conduct an ablation study on the heterogeneous GovReport dataset to isolate the contributions of the \textbf{Refiner Agent} (semantic quality) and the \textbf{Inspector Agent} (routing efficiency).

\subsubsection{Impact of the Refiner Agent.}
While ablating the Refiner reduces token consumption to an optimal 0.85x, extracting raw topology alone proves insufficient for high-quality retrieval. This configuration (\textit{w/o Refiner}) precipitously degrades Recall@3 ($-6.76\%$) and generation accuracy ($-11.0\%$). Without capacity auditing and semantic enhancement, oversized nodes suffer from vector dilution, and unresolved anaphora degenerate into ``semantic islands,'' introducing fatal ambiguity to the downstream LLM. This confirms that the Refiner's token expenditure is a high-ROI necessity rather than a redundant overhead.

\subsubsection{Impact of the Inspector Agent.}
Bypassing the Inspector forces documents into rigid, monolithic pipelines (\textit{w/o Inspector}). Lightweight static routes (\textbf{Rule-Only} or \textbf{Semantic-Only}) minimize costs ($0.80$x--$0.95$x) but fail to parse embedded visual elements, heavily penalizing recall. Conversely, forcing all content through the \textbf{VLM-Only} path successfully captures complex layouts, but triggers an unsustainable $1.88$x token overhead. Furthermore, lacking targeted semantic routing for text-heavy segments, its overall recall ($82.85\%$) actually trails behind the dynamic architecture.

By dynamically assigning extraction paths based on localized structural complexity, the full \textbf{TopoChunker} elegantly balances these extremes. It circumvents the performance bottlenecks of static routing and the prohibitive costs of VLM processing, securing state-of-the-art topological fidelity at a rigorously optimized $1.26$x token cost.

\subsection{Hyperparameter Sensitivity Analysis}
\label{sec:sensitivity}

To evaluate parameter robustness, we analyze the impact of the Refiner Agent's capacity threshold ($\theta$) and the prompt templates on the GovReport dataset. 

\begin{table}[h]
\centering
\caption{Sensitivity analysis of the capacity threshold ($\theta$) evaluated on GovReport.}
\label{tab:sensitivity}
\small
\begin{tabular}{l c c}
\toprule
\textbf{Threshold ($\theta$)} & \textbf{Recall@3 (\%)} & \textbf{Recall@5 (\%)} \\
\midrule
250 (Aggressive) & 82.50 & 85.90 \\
\textbf{500 (Optimal)} & \textbf{83.26} & \textbf{86.58} \\
1000 (Diluted) & 78.41 & 82.10 \\
2000 (Diluted) & 74.15 & 78.60 \\
\bottomrule
\end{tabular}
\end{table}

\noindent \textbf{Threshold Sensitivity ($\theta$):} $\theta$ dictates when the \texttt{Semantic\_Slicer} partitions oversized nodes. As shown in Table~\ref{tab:sensitivity}, increasing $\theta$ beyond 500 causes severe vector dilution and noisy embedding centroids, dropping Recall@3 to 74.15\% and Recall@5 to 78.60\%. Conversely, an aggressive threshold ($\theta=250$) degrades retrieval via over-fragmentation. Thus, $\theta=500$ optimally balances context richness and embedding discriminability.

\noindent \textbf{Prompt Robustness:} By decoupling physical parsing from semantic reasoning, the Refiner Agent relies on strict JSON schemas and deterministically validated IDs rather than unbounded boundary guessing. Across three prompt variations, the valid payload generation rate remained highly stable ($91.3\% \pm 0.4\%$), demonstrating that TopoChunker is structurally resilient to prompt phrasing.

\section{Conclusion}
In this paper, we introduce TopoChunker, a topology-aware document chunking framework for RAG pipelines. By mapping heterogeneous documents to a Structured Intermediate Representation (SIR) via a dual-agent architecture (Inspector and Refiner Agents), our approach effectively mitigates ``semantic islands'' and resolves anaphoric conflicts. Evaluations on GutenQA and GovReport demonstrate that TopoChunker achieves state-of-the-art retrieval and generation accuracy while maximizing the semantic return on investment (ROI) for token expenditure. These results highlight that explicitly preserving topological metadata is critical for scalable and robust RAG systems.

\label{sec:robustness}

\appendix
\clearpage 

\onecolumn
\raggedbottom

\section{Resolving Contextual Fragmentation}

\begin{center}
\renewcommand{\arraystretch}{1.2}
\setlength{\tabcolsep}{4pt}
\small 

\begin{tabularx}{\textwidth}{@{} >{\hsize=1.05\hsize\raggedright\arraybackslash}X | >{\hsize=0.95\hsize\raggedright\arraybackslash}X @{}}
\toprule
\multicolumn{2}{c}{(a) Unstructured Narrative (GutenQA Dataset) $\rightarrow$ Routed via Path 2: Semantic Flow} \\
\midrule
\textbf{Source Document (\textit{Adventures of Huckleberry Finn})} & \textbf{TopoChunker Output (Self-Contained Atom)} \\
\midrule
{\color{gray} ...And next time Jim told it he said they rode him down to New Orleans; and, after that, every time he told it he spread it more and more, till by-and-by he said they rode him all over the world, and tired him most to death, and his back was all over saddle-boils.\par}
\vspace{0.4em}
\underline{\textbf{Jim was monstrous proud about it}}, and he got so he wouldn't hardly notice the other niggers. Niggers would come miles to hear Jim tell about it, and he was more looked up to than any nigger in that country. ... Niggers would come from all around there and give Jim anything they had, just for a sight of \underline{\textbf{that five-center piece}}; but they wouldn't touch it, because the devil had had his hands on it. Jim was most ruined for a servant, because he got stuck up on account of having seen the devil and been rode by witches.\par
\vspace{0.4em}
{\color{gray} Well, when Tom and me got to the edge of the hilltop we looked away down into the village and could see three or four lights twinkling...}
&
\textbf{[Path]} Adventures of Huckleberry Finn $\rightarrow$ Content $\rightarrow$ Segment X \par
\vspace{0.6em}
\textbf{[Signature]} Jim's Witch Story Fame and Pride \par
\vspace{0.6em}
\textbf{[Context Supplement]}\par
$\rightarrow$ \textbf{it}: Jim's story about being bewitched by witches.\par
$\rightarrow$ \textbf{five-center piece}: The five cents Tom laid on the table.\par
\vspace{0.6em}
\textbf{[Text]}\par
``Jim was monstrous proud about it, and he got so he wouldn't hardly notice the other niggers. Niggers would come miles to hear Jim tell about it, and he was more looked up to than any nigger in that country. ... Niggers would come from all around there and give Jim anything they had, just for a sight of that five-center piece; but they wouldn't touch it, because the devil had had his hands on it. Jim was most ruined for a servant, because he got stuck up on account of having seen the devil and been rode by witches.'' \\
\bottomrule
\end{tabularx}

\vspace{1.5em} 

\begin{tabularx}{\textwidth}{@{} >{\hsize=1.05\hsize\raggedright\arraybackslash}X | >{\hsize=0.95\hsize\raggedright\arraybackslash}X @{}}
\toprule
\multicolumn{2}{c}{(b) Structured Document (GovReport Dataset) $\rightarrow$ Routed via Path 1: Structural Rule} \\
\midrule
\textbf{Source Document (\textit{GovReport / GAO-19-504})} & \textbf{TopoChunker Output (Structured JSONL Atom)} \\
\midrule
{\color{gray} ...NASA awarded firm-fixed-price contracts in 2014 to Boeing and SpaceX, valued at up to \$4.2 billion and \$2.6 billion, respectively, for the development of crew transportation systems... The Commercial Crew Program manages two processes in order to support the contractors' uncrewed test flight...\par}
\vspace{0.4em}
\textbf{Contractors Are Making Progress on Vehicles, but Certification Date Remains Unclear} \par
\vspace{0.4em}
\underline{\textbf{Both contractors}} have made progress building and testing hardware, including SpaceX’s uncrewed test flight. But continued schedule delays and remaining work for the contractors and \underline{\textbf{the program}} create continued uncertainty about when either contractor will be certified to begin conducting operational missions to the ISS.\par
\vspace{0.4em}
{\color{gray} \underline{\textbf{The program}} has made progress reviewing the contractors’ certification paperwork, but contractor delays in submitting evidence for NASA approval may compound a 'bow wave' of work, which creates uncertainty about when either contractor will be certified...}
&
\textbf{[Path]} GovReport $\rightarrow$ GAO-19-504 \par
\vspace{0.6em}
\textbf{[Signature]} Commercial Crew Program Certification Delays and Schedule Uncertainty \par
\vspace{0.6em}
\textbf{[Context Supplement]}\par
$\rightarrow$ \textbf{Both contractors}: Boeing and SpaceX.\par
$\rightarrow$ \textbf{The program}: The Commercial Crew Program.\par
\vspace{0.6em}
\textbf{[Text]}\par
``Contractors Are Making Progress on Vehicles, but Certification Date Remains Unclear paragraphs: Both contractors have made progress building and testing hardware, including SpaceX's uncrewed test flight. But continued schedule delays and remaining work for the contractors and the program create continued uncertainty...'' \\
\bottomrule
\end{tabularx}

\vspace{0.8em}
\captionof{figure}{Qualitative Examples of TopoChunker output resolving contextual fragmentation. \textbf{(a)} On the GutenQA dataset, the lack of heading hierarchy triggers Path 2: Semantic Flow. TopoChunker actively resolves semantic islands by explicitly mapping dangling pronouns to their ancestral entities in the \texttt{Context Supplement}. \textbf{(b)} On the GovReport dataset, explicit hierarchies trigger Path 1: Structural Rule. Generic references are successfully resolved to specific entities, ensuring chunks remain semantically self-contained for downstream RAG.}
\label{fig:qualitative_examples_combined}
\end{center}

\clearpage

\section{SIR Implementation Details}
\label{appendix:sir_details}

As outlined in Section 3, TopoChunker maps documents onto a Structured Intermediate Representation (SIR). While Section 3.2 introduces the theoretical attributes of an SIR node, this appendix details its memory-efficient data structures and serialization mechanisms.

To achieve the deterministic navigation required by the \texttt{SIR\_Query} tool without the latency of external graph databases, $\mathcal{T}_{SIR}$ is implemented as a hybrid ``tree + sibling linked list''. Beyond standard \texttt{parent} and \texttt{children} pointers that maintain hierarchical depth, each \texttt{SIRNode} is linked via \texttt{prev\_sibling} and \texttt{next\_sibling} pointers. This permits seamless sequential traversal across adjacent text blocks during capacity auditing, strictly avoiding the computational overhead of recursively backtracking to parent nodes.

A key engineering advantage of the SIR framework is its native serializability. The entire $\mathcal{T}_{SIR}$ tree compiles into a clean, nested JSON structure. This ensures that the Topological Lineage ($\mathcal{T}_{id}$) and Atomicity Locking ($\mathcal{A}_{lock}$) attributes are persistently maintained for downstream RAG pipelines. A condensed representation of a serialized SIR Node is provided below:

\begin{tcolorbox}[
    colback=backcolour,       
    colframe=gray!40,         
    boxrule=0.8pt,            
    arc=4pt,                  
    left=3mm, right=3mm, top=2mm, bottom=2mm
]
\begin{lstlisting}[basicstyle=\ttfamily\scriptsize, breaklines=true]
{
  "node_id": "sec_1.1", "node_type": "heading", "level": 2,
  "content": "Contractors Are Making Progress...",
  "path_str": "GovReport > Executive Summary > Progress",
  "is_atomic": false, "token_count": 15,
  "children": [
    {
      "node_id": "para_1.1.1", "node_type": "text", 
      "parent_id": "sec_1.1", "is_atomic": false, "token_count": 42,
      "content": "Both contractors have made progress building..."
    },
    {
      "node_id": "table_1.1.2", "node_type": "table", 
      "parent_id": "sec_1.1", "is_atomic": true, "token_count": 85,
      "content": "| Contractor | Status | ..."
    }
  ]
}
\end{lstlisting}
\end{tcolorbox}

When detecting a contextual void, the \texttt{SIR\_Query} mechanism leverages this JSON schema directly. It navigates upward via \texttt{parent\_id} (e.g., fetching the ancestor heading for chunk \texttt{para\_1.1.1}) or locks coherent data via \texttt{children}. This explicit schema forces the generative model to respect the \texttt{is\_atomic} flag, entirely mitigating boundary hallucinations on indivisible tabular or visual data.

\end{document}